\documentclass{article}
\usepackage{ijcai22}

\usepackage{times}
\usepackage{soul}
\usepackage{url}
\usepackage[hidelinks]{hyperref}
\usepackage[utf8]{inputenc}
\usepackage[small]{caption}
\usepackage{graphicx}
\usepackage{amsmath}
\usepackage{amsthm}
\usepackage{booktabs}
\usepackage{algorithm}
\usepackage{algorithmic}
\usepackage{multirow}
\usepackage{diagbox}

\title{Improving the Transferability of Adversarial Examples 
\\
with Restructure Embedded Patches}

\author{
Huipeng Zhou$^1$\and
Yu-an Tan$^2$\and
Yajie Wang$^2$\and
Haoran Lyu$^2$\and
Shangbo Wu$^2$\\ \And
Yuanzhang Li\thanks{Corresponding author.}$^{,1}$
\affiliations
$^1$School of Computer Science and Technology, Beijing Institute of Technology\\
$^2$School of Cyberspace Science and Technology, Beijing Institute of Technology
\emails
\{zhouhuipeng, popular\}@bit.edu.cn
}

\begin{document}

\maketitle

\begin{abstract}
Vision transformers (ViTs) have demonstrated impressive performance in various computer vision tasks. However, the adversarial examples generated by ViTs are challenging to transfer to other networks with different structures. Recent attack methods do not consider the specificity of ViTs architecture and self-attention mechanism, which leads to poor transferability of the generated adversarial samples by ViTs. We attack the unique self-attention mechanism in ViTs by restructuring the embedded patches of the input. The restructured embedded patches enable the self-attention mechanism to obtain more diverse patches connections and help ViTs keep regions of interest on the object. Therefore, we propose an attack method against the unique self-attention mechanism in ViTs, called Self-Attention Patches Restructure (SAPR). Our method is simple to implement yet efficient and applicable to any self-attention based network and gradient transferability-based attack methods. We evaluate attack transferability on black-box models with different structures. The result show that our method generates adversarial examples on white-box ViTs with higher transferability and higher image quality. Our research advances the development of black-box transfer attacks on ViTs and demonstrates the feasibility of using white-box ViTs to attack other black-box models.
\end{abstract}

\begin{figure}[t]
    \centering
    \includegraphics[width=\linewidth]{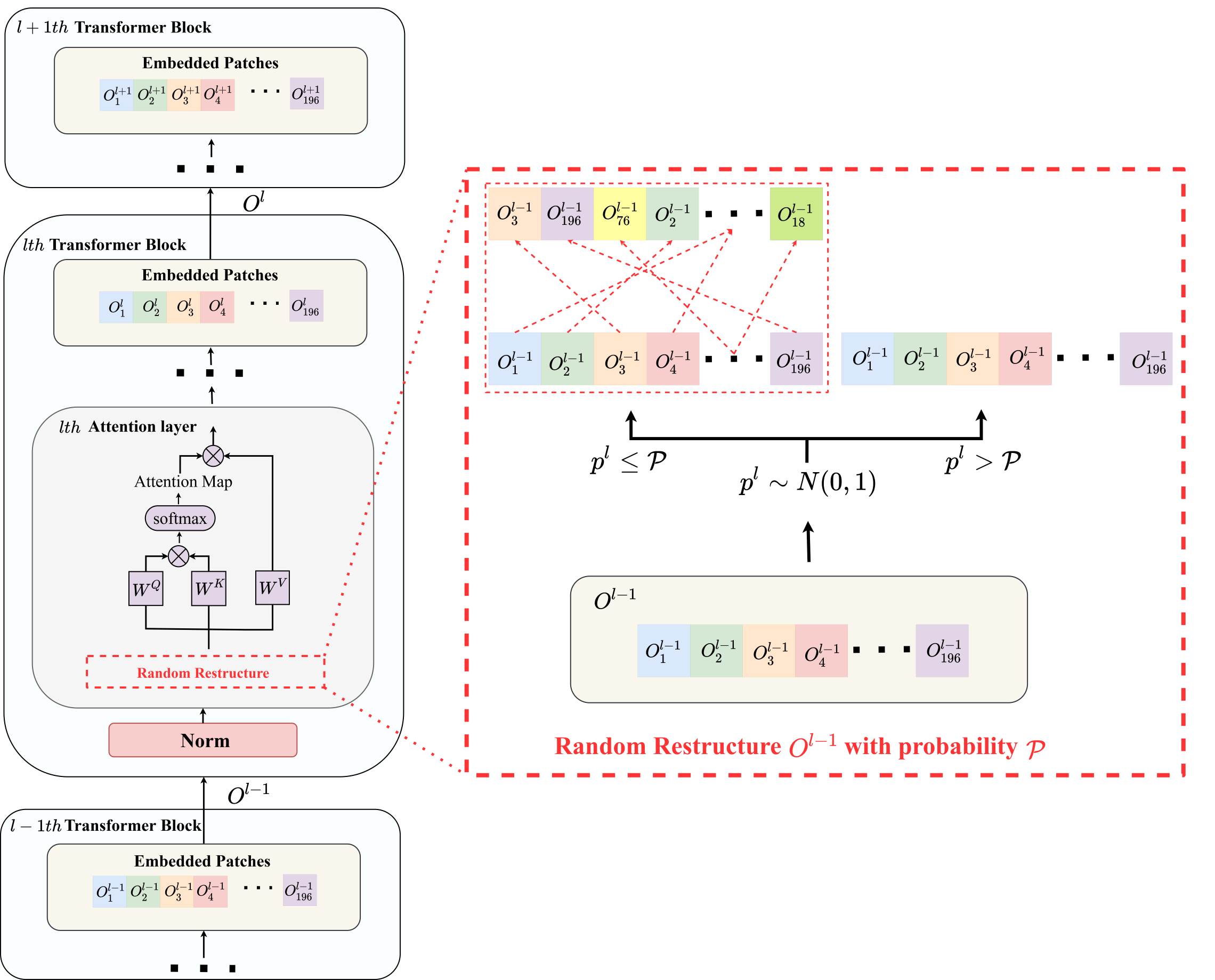}
    \caption{Self-Attention Patches Restructure (SAPR) method.  For the attention layer in the $l$th transformer block accept the output embedded patches $O^{l-1}$ from the $l-1$th block. $p^l$ is the probability defined in the $l$th attention layer. The $p^l$ is sampled during forwarding propagation, and only if $p^l\le\mathcal{P}$ then patches are restructured for $O^{l-1}$ and the original patches alignment distribution is disrupted.}\label{fig:1}
\end{figure}

\section{Introduction}
Vision Transformers (ViTs)~\cite{Dosovitskiy2021AnII,Touvron2021TrainingDI} have achieved satisfying performance on computer vision tasks. ViTs have led to a new wave of research in computer vision because of their unique network architecture and different image processing approaches compared to CNNs. Nevertheless, ViTs are still vulnerable to security threats from adversarial examples. Adversarial examples are malicious images with imperceptible perturbations, leading deep neural networks to misclassification of high confidence.

The adversarial samples have the property of being transferable cross-model ~\cite{Xie2019ImprovingTO,Dong2018BoostingAA}, making it possible to attack unknown black-box models without knowing the architecture and properties ~\cite{Liu2017DelvingIT}. Many well-established gradient transferability-based black-box attack methods on CNNs typically use data augmentation and advanced gradient computation to generate adversarial examples. In contrast, little is known about black-box attack methods on ViTs. The differences in structure, parameters, and image processing between ViTs and other models lead to poor transferability of the generated adversarial examples. Because the unique architecture of ViTs is not considered, extending methods that perform well on existing CNNs to ViTs is not easy and has unsatisfactory results. Recent work ~\cite{Naseer2021OnIA} proposes a self-Ensemble (SE) method, which uses class tokens at each layer and a shared classification head to construct a model ensemble. Although this method can improve the transferability of adversarial examples generated on ViTs, the improvement is not significant. It does not consider the different structures of black-box models and the unique self-attention mechanism of ViTs.

ViTs takes as input a flattened sequence of patches from an image and uses a series of multi-headed self-attention (MSA) layers to learn the connections between the patches ~\cite{Vaswani2017AttentionIA}. The self-attention mechanism can obtain the connections between patches and thus get the information of global images. Recent attack methods do not consider the specificity of ViTs architecture and self-attention mechanism, which leads to poor transferability of the generated adversarial samples. In this work, we attack the unique self-attention mechanism in ViTs, breaking the connection between patches in self-attention. We break the connections between the original patches by restructuring the input embedded patches in the self-attention layer with probability $\mathcal{P}$. The restructured embedded patches enable the self-attention mechanism to obtain more diverse patches connections, avoid over-fitting some patches during adversarial attacks, and consider the impact of global patches on transferability. At the same time, the restructured embedded patches can help ViTs keep regions of interest on objects during adversarial attacks and generate higher quality adversarial samples. To this end, we propose an attack method against the unique self-attention mechanism in ViTs, called Self-Attention Patch Restructure (SAPR), as shown in Fig.~\ref{fig:1}. Our method is simple to implement yet efficient and can be easily combined with existing attack methods, and is suitable for any self-attention-based network.

We evaluate the transferability of attacks on state-of-the-art ViTs, CNNs and MLPs models. The experimental results confirm that our proposed method can improve the transferability of adversarial examples generated by ViTs on different structural black-box models and improve the performance of existing gradient-based transferability attack methods on ViTs.  To the best of our knowledge, our method improves the average black-box attack success rate by 9.71\% compared to the current state-of-the-art SE methods. When combined with SE methods, we can improve the average black-box attack success rate by 19.35\% compared to SE methods. Our method enhances the transferability and the quality of the adversarial examples. Our research advances the development of black-box migration attacks on ViTs and demonstrates the feasibility of using white-box ViTs to attack other black-box models.

We briefly summarize our primary contributions as follows:
\begin{itemize}
    \item We attack the unique self-attention mechanism in ViTs, breaking the connection between the original patches in self-attention. We improve the transferability of adversarial examples generated by ViTs against black-box models with different structures.
    \item We propose an attack method against the unique self-attention mechanism in ViTs, called Self-Attention Patch Restructure (SAPR). It is simple to implement yet efficient and can be easily combined with existing attack methods and improve their performance on ViTs.
    \item We evaluate the transferability of attacks on state-of-the-art ViTs, CNNs and MLPs models. Experiments confirm that our method generates adversarial examples by ViTs with higher transferability and higher image quality. 
\end{itemize}

\section{Related Work}
 \quad\textbf{Vision Transformers.}The Vision Transformers (ViTs) was first proposed by ~\cite{Dosovitskiy2021AnII}. ViT takes the image patches as input and pre-trains them using a huge dataset. To overcome the factor of model pre-training ViTs, the massive dataset-based DeiT ~\cite{Touvron2021TrainingDI} introduced a transformer-specific teacher-student strategy that uses a new distillation token to learn knowledge from CNNs. T2T- vit ~\cite{Yuan2021TokenstoTokenVT} introduced the T2T module to model the local structure of an image and uses the deep-narrow structure as the backbone of the transformer. Swin ~\cite{Liu2021SwinTV} allows the model to learn information across the window by introducing a sliding window mechanism.

\textbf{Black-box attack.}Transferability-based attacks on CNNs have been well studied and have achieved satisfactory attack success rates. The momentum iteration (MI) attack ~\cite{Dong2018BoostingAA} is designed to stabilize the update direction by integrating momentum terms and avoiding local optima. The diversity input (DI) attack ~\cite{Xie2019ImprovingTO} randomly resizes and fills the input with a fixed probability at each iteration. The scale-invariant method (SIM) ~\cite{Lin2020NesterovAG} optimizes the ingestion of the scaled image ensemble at each iteration.

Compared to transfer-based attacks on CNNs, less effort has been made to investigate the transferability of adversarial examples between white-box ViTs and different structural black-box models. A related work ~\cite{Naseer2021OnIA} proposes a self-Ensemble (SE) method, which uses class labelling at each layer and a shared classification head to construct a model ensemble. The transferability of adversarial examples is improved by optimizing the perturbations on the model ensemble. In addition, ~\cite{Naseer2021OnIA} introduces a token refine (TR) module to fine-tune the class tokens to enhance transferability. Although this method can improve the transferability of adversarial examples generated on ViTs, the improvement is not significant. It does not consider the different structures of black-box models and the unique self-attention mechanism of ViTs. In addition, TR requires access to the ImageNet training set during fine-tuning, which is very costly and timely. In contrast, our approach can be generalized to different ViTs models without any external auxiliary conditions and is easy to implement.

\section{Methodology}
Given a classifier $f(x):x \in X \to y \in Y$, it outputs a label $y$ as a prediction of the input $x$. The goal of the adversarial attack is to find an example $x^{*}$ near $x$ that mistakes the classifier for $f(x^{*}) \neq y$. $x^{*}$ and $x$ satisfy $\parallel x^{*}-x \parallel_{p} <\epsilon$, for a small perturbation budget $\epsilon$ and a p-norm that is usually set to an $l_{\infty}$-norm.

ViTs have noticeable structural differences from CNNs. Specifically, ViTs take as input a flattened sequence of patches from an image and use a series of multi-headed self-attention (MSA) layers to learn the connections between patches. The self-attention mechanism can obtain the connections between patches and thus get the information of global images. In contrast, in CNNs, image information is extracted by convolutional kernels in the convolutional layer but is local to the image. Extending well-established attack methods to ViTs does not consider the specificity of its architecture and self-attention mechanism, resulting in poor performance.

There is a unique structure of multi-headed self-attention in ViTs. By projecting concatenated outputs from multiple heads, MSA combines information from different representation subspaces ~\cite{Vaswani2017AttentionIA}. There are three significant vectors in the Attention structure: Query (to match others), Key (to be checked), and V (information to be extracted) usually expressed as Q, K, V, which are taken from the output matrix of the learned Linear layer and then mapped to the relative dimensions.

There are $n$ transformer blocks in the ViTs model, and each block has an attention layer. Take the $l$th block as an example. The $l$th attention layer accepts the output $O^{l-1} \in R(d, N)$ of the embedded patches of the $l-1$th block as the input matrix. Multiply $O^{l-1} \in R(d, N)$ with the weights $W$ in the Linear layer of the attention layer to get the intermediate matrices $Q, K, V \in R(d, N)$. Transpose $K$ and multiply it with $Q$ to get attention map $A \in R(N, N)$, in which each position represents the connection between two patches. Then take the attention map by softmax operation to get the attention map $A^\ast \in R(N, N)$, and finally multiply it with the $V$ matrix to get the final output matrix $O^l \in R(d, N)$. We also consider the output matrix-vector $O$ as the feature vector in the attention layer. Thus the feature vector in the $l$th attention layer in ViTs can be expressed using the following equations:
\begin{equation}\label{con:1} 
    Q,K,V = \mathrm{Linear}(O^{l-1})
\end{equation}
\begin{equation}\label{con:2}   
    O^l=\mathrm{Attention}(Q,K,V)=\mathrm{softmax}(\frac{Q \bullet K^{T}}{\sqrt{d_k}})V
\end{equation}

All the image information is stored in the self-attention mechanism, and $A$ stores the connections between different patches. We get different $Q$, $K$, $V$ by restructuring the input matrix $O^{l-1}$ with probability $\mathcal{P}$ to break the connection between patches and attack the self-attention mechanism. The restructured embedded patches allow A to obtain connections with diverse patches. Diverse A avoids the problem of overfitting in some patches and considers the effect of global patches on the transferability of adversarial examples. When adversarial attacks are performed, the restructured embedded patches cause the self-attention mechanism to focus on the object. The generated adversarial perturbations focus on the object and generate adversarial examples with higher transferability and higher quality. Given a ViT model $F$, the clean image $x_{ori}$ with the ground-truth label as $y$. The embedded patches restructuring probability threshold $\mathcal{P}$. We want to find an example $x_{adv}$ near $x_{ori}$ mistakes the classifier as $F(x_{adv}) \neq y$. ViTs model has $n$ attention layers, and we define the probability $p \sim N(0,1)$ in each attention layer. As Fig.\ref{fig:1}. For the $l$th attention layer, $p^l \sim N(0,1) $will be resampled at each forward propagation. If $p^l \le \mathcal{P}$ then random restructuring is performed on the embedded patches input of the $l$th attention layer. If $p^l > \mathcal{P}$ then no operation is done. For a ViT model, the probability $p$ in all the attention layers inside it can be regarded as a set, that is $p=\{p^1,p^2,p^3,\cdots,p^n\}$. Therefore, the feature vector in the $l$th attention layer in ViTs can be expressed by the following equations:

\begin{equation}
\begin{aligned}\label{con:3} 
    Q^{'},K^{'},V^{'} = \mathrm{Linear}(\mathrm{Restructure}(O^{l-1},p^l))   
    \\
    \mathrm{when} \quad p^l \le \mathcal{P} \quad \mathrm{do} \quad \mathrm{Restructure}
\end{aligned}\end{equation}

\begin{equation}\label{con:4} 
     O^l=\mathrm{Attention}(Q^{'},K^{'},V^{'})=\mathrm{softmax}(\frac{Q^{'} \bullet (K^{'})^{T}}{\sqrt{d_k}})V^{'}
\end{equation}
Where $Restructure$ denotes the random restructuring function. Since the input to ViTs is a flattened patch sequence, it is only necessary to disrupt the whole patch arrangement and perform random combinations. Thus the optimization problem in solving the adversarial example becomes the following equation:
\begin{equation}\label{con:5} 
    \mathrm{argmax}\mathrm{J}(F(x_{ori}+\delta,\mathcal{P}),y),s.t \parallel \delta \parallel_{\infty} <\epsilon
\end{equation}
where $\mathrm{J}$ denotes the loss function is often chosen as the cross-entropy loss, and $\delta$ is the perturbation. Our proposed method can be combined with any gradient transferability based method, and our attack method is shown in Algorithm ~\ref{algorithm:1}.

\begin{algorithm}[tb]
    \caption{SAPR attack}
    \label{alg:algorithm}
    \textbf{Input}: Original image $x$ with ground-truth label y, perturbation budget of $l_{p}$-normd $\epsilon$, step size $\alpha$.\\
    \textbf{Parameter}: ViTs Classification Model $F$\ with loss function $J$, attack iterations $T$, Attention layer patches random restructuring threshold $\mathcal{P}$, number of Attention layers $n$. Attention layer internal probability $p=\{p^1,p^2,p^3,\cdots,p^n\}$\\
    \textbf{Output}: adversarial example $x_{adv}.$
    \begin{algorithmic}[1] 
    \renewcommand{\baselinestretch}{2}
        \STATE $x_{adv} \gets x$.
        \WHILE{$i<T$}
            \STATE $p \gets p^{1-n} \sim N(0,1)$ // in each attention layer
            \STATE $output = F(x_{adv},\mathcal{P})$.
            \STATE $g \gets \bigtriangledown_{\delta}J(output,y)$
            \STATE $x_{adv} \gets clip_{x,\epsilon}(x_{adv}+\alpha \cdot g)$
        \ENDWHILE
        \STATE \textbf{return} $x_{adv}$
    \end{algorithmic}
    \label{algorithm:1}
\end{algorithm}

\begin{table*}[t]
\centering
\setlength{\tabcolsep}{1mm}{
\scalebox{0.9}{
\begin{tabular}{@{}c|c|c|ccccccccc|c@{}}
\toprule
model & $\mathcal{P}$ & method & VGG16 & ResNet50 & \begin{tabular}[c]{@{}c@{}}RepVGG\\ -A0\end{tabular} & \begin{tabular}[c]{@{}c@{}}AdvEffici\\ entNetb0\end{tabular} & \begin{tabular}[c]{@{}c@{}}T2T-\\ vit-19\end{tabular} & \begin{tabular}[c]{@{}c@{}}ViT-\\ L/32\end{tabular} & \begin{tabular}[c]{@{}c@{}}DeiT\\ -tiny\end{tabular} & \begin{tabular}[c]{@{}c@{}}Mixer-\\ B/16\end{tabular} & \begin{tabular}[c]{@{}c@{}}SwinMLP\\ -T/C6\end{tabular} & average \\ \midrule
\multirow{6}{*}{T2T-vit-24} & \multirow{6}{*}{0.5} & MIM & 31.60\% & 20.10\% & 27.60\% & 12.70\% & 74.50\% & 5.70\% & 14.50\% & 13.00\% & 17.00\% & 24.08\% \\
 &  & MIM+Ours & \textbf{64.70\%} & \textbf{40.90\%} & \textbf{53.20\%} & \textbf{23.60\%} & \textbf{97.40\%} & \textbf{8.00\%} & \textbf{39.70\%} & \textbf{24.60\%} & \textbf{36.00\%} & \textbf{43.12\%} \\
 &  & DIM & 33.10\% & 25.40\% & 33.30\% & 12.50\% & 92.50\% & 8.10\% & 21.20\% & 18.30\% & 26.70\% & 30.12\% \\
 &  & DIM+Ours & \textbf{74.30\%} & \textbf{57.70\%} & \textbf{70.90\%} & \textbf{31.70\%} & \textbf{99.00\%} & \textbf{11.80\%} & \textbf{57.40\%} & \textbf{34.70\%} & \textbf{57.70\%} & \textbf{55.02\%} \\
 &  & SIM & 35.40\% & 25.50\% & 32.10\% & 14.80\% & 88.40\% & 7.00\% & 19.90\% & 17.80\% & 20.30\% & 29.02\% \\
 &  & SIM+Ours & \textbf{69.90\%} & \textbf{54.70\%} & \textbf{66.20\%} & \textbf{31.60\%} & \textbf{99.30\%} & \textbf{17.00\%} & \textbf{57.20\%} & \textbf{39.80\%} & \textbf{56.40\%} & \textbf{54.68\%} \\ \midrule
\multirow{6}{*}{ViT-B/16} & \multirow{6}{*}{0.15} & MIM & 22.20\% & 12.20\% & 19.20\% & 10.10\% & 6.00\% & 55.00\% & 32.20\% & 15.40\% & 7.90\% & 20.02\% \\
 &  & MIM+Ours & \textbf{31.00\%} & \textbf{19.30\%} & \textbf{27.80\%} & \textbf{16.20\%} & \textbf{14.80\%} & \textbf{92.80\%} & \textbf{70.50\%} & \textbf{33.10\%} & \textbf{14.30\%} & \textbf{35.53\%} \\
 &  & DIM & 13.50\% & 10.50\% & 15.80\% & 8.60\% & 9.00\% & 75.10\% & 38.70\% & 17.20\% & 9.90\% & 22.03\% \\
 &  & DIM+Ours & \textbf{23.30\%} & \textbf{18.30\%} & \textbf{27.00\%} & \textbf{14.70\%} & \textbf{18.50\%} & \textbf{94.10\%} & \textbf{69.30\%} & \textbf{32.40\%} & \textbf{17.90\%} & \textbf{35.06\%} \\
 &  & SIM & 28.00\% & 14.80\% & 22.30\% & 13.20\% & 9.00\% & 69.90\% & 41.90\% & 20.70\% & 12.10\% & 25.77\% \\
 &  & SIM+Ours & \textbf{36.10\%} & \textbf{28.80\%} & \textbf{38.70\%} & \textbf{24.20\%} & \textbf{34.10\%} & \textbf{98.00\%} & \textbf{85.70\%} & \textbf{51.20\%} & \textbf{29.40\%} & \textbf{47.36\%} \\ \midrule
\multirow{6}{*}{DeiT-small} & \multirow{6}{*}{0.3} & MIM & 32.90\% & 18.80\% & 25.60\% & 14.20\% & 13.10\% & 21.80\% & 76.40\% & 30.10\% & 13.20\% & 27.34\% \\
 &  & MIM+Ours & \textbf{55.10\%} & \textbf{36.90\%} & \textbf{52.20\%} & \textbf{30.30\%} & \textbf{34.50\%} & \textbf{60.80\%} & \textbf{99.50\%} & \textbf{75.30\%} & \textbf{33.30\%} & \textbf{53.10\%} \\
 &  & DIM & 32.00\% & 24.90\% & 33.30\% & 19.10\% & 34.10\% & 37.20\% & 91.20\% & 45.80\% & 28.80\% & 38.49\% \\
 &  & DIM+Ours & \textbf{52.80\%} & \textbf{39.60\%} & \textbf{54.70\%} & \textbf{32.50\%} & \textbf{43.90\%} & \textbf{60.10\%} & \textbf{98.20\%} & \textbf{71.10\%} & \textbf{42.80\%} & \textbf{55.08\%} \\
 &  & SIM & 36.40\% & 22.10\% & 31.80\% & 19.30\% & 21.40\% & 33.10\% & 82.20\% & 40.20\% & 17.80\% & 33.81\% \\
 &  & SIM+Ours & \textbf{65.40\%} & \textbf{52.40\%} & \textbf{66.60\%} & \textbf{42.90\%} & \textbf{59.10\%} & \textbf{80.80\%} & \textbf{99.90\%} & \textbf{88.90\%} & \textbf{55.40\%} & \textbf{67.93\%} \\ \midrule
\multirow{6}{*}{Swin-L} & \multirow{6}{*}{0.3} & MIM & 23.70\% & 9.10\% & 14.20\% & 6.00\% & 6.50\% & 3.80\% & 5.50\% & 8.10\% & 13.50\% & 10.04\% \\
 &  &MIM+Ours & \textbf{42.80\%} & \textbf{20.60\%} & \textbf{26.00\%} & \textbf{10.20\%} & \textbf{22.00\%} & \textbf{6.10\%} & \textbf{18.50\%} & \textbf{18.50\%} & \textbf{34.30\%} & \textbf{22.11\%} \\
 &  & DIM & 22.60\% & 11.30\% & 14.60\% & 4.90\% & 12.30\% & 4.30\% & 5.50\% & 7.40\% & 21.50\% & 11.60\% \\
 &  & DIM+Ours & \textbf{44.50\%} & \textbf{27.50\%} & \textbf{31.30\%} & \textbf{12.50\%} & \textbf{32.90\%} & \textbf{8.00\%} & \textbf{21.30\%} & \textbf{21.70\%} & \textbf{50.10\%} & \textbf{27.76\%} \\
 &  & SIM & 25.10\% & 9.80\% & 14.40\% & 6.40\% & 9.70\% & 4.60\% & 7.00\% & 7.60\% & 15.60\% & 11.13\% \\
 &  & SIM+Ours & \textbf{46.20\%} & \textbf{29.10\%} & \textbf{33.10\%} & \textbf{15.10\%} & \textbf{37.40\%} & \textbf{10.60\%} & \textbf{28.40\%} & \textbf{28.70\%} & \textbf{52.60\%} & \textbf{31.24\%} \\ \bottomrule
\end{tabular}
}}
\caption{The table reports the attack success rate of 1000 adversarial examples generated by white-box ViTs on different structural black-box models. Our method can improve the average attack success rate of existing methods on black-box ViTs by \{MIM: 20.08\%, DIM: 15.44\%, SIM: 26.12\%\} and on black-box CNNs, robust CNNs and MLPs by \{MIM: 16.75\%, DIM: 18.78\%, SIM: 25.00\%\}.}\label{table:1}
\end{table*}

\section{Experiments and Results}
\paragraph{Experimental Settings.}
In this section, we perform experiments on many ViTs, CNNs, robust CNNs and MLPs. We selected 11 ViTs models as white-box models. They are T2T-vit-24, T2T-vit-19, ViT-B/16, ViT-L/32, DeiT-tiny, DeiT-small, DeiT-B/16, DeiT-distilled-small, Swin-B ~\cite{Liu2021SwinTV}, Swin-L, Swin-T. We perform transferability experiments on black-box models with different structures. The chosen model types are DenseNet ~\cite{Huang2017DenselyCC}, VGG ~\cite{Simonyan2015VeryDC}, ResNet ~\cite{He2016DeepRL}, Inception ~\cite{Szegedy2015GoingDW}, MobileNet ~\cite{Howard2017MobileNetsEC}, WideResNet ~\cite{Zagoruyko2016WideRN}, Efficientnet ~\cite{Tan2019EfficientNetRM}, T2T, ViT, DeiT, MixerMLP ~\cite{Tolstikhin2021MLPMixerAA},  RepVGG ~\cite{Ding2021RepVGGMV}, ReXNet ~\cite{Han2021RethinkingCD}, Swin and GhostNet ~\cite{Han2020GhostNetMF}. Specific details can be found in the supplement. We chose 1000 images from Imagenet ~\cite{Russakovsky2015ImageNetLS} for our experiments, and all of them can be successfully classified by the model we chose.

We use currently accepted gradient transferability-based attack methods in combination with our method, namely MI-FGSM (MIM), DI-FGSM (DIM), and SI-FGSM (SIM). These methods are often used as components by other attack methods. We use the $\ell_\infty$-norm with $\epsilon$=16 for the constraintson the adversarial examples.  The number of iteration roundsis chosen to be 50.

\begin{figure*}[htbp]
    \centering
    \begin{minipage}[t]{0.33\textwidth} 
			\centering
			\includegraphics[width=\textwidth]{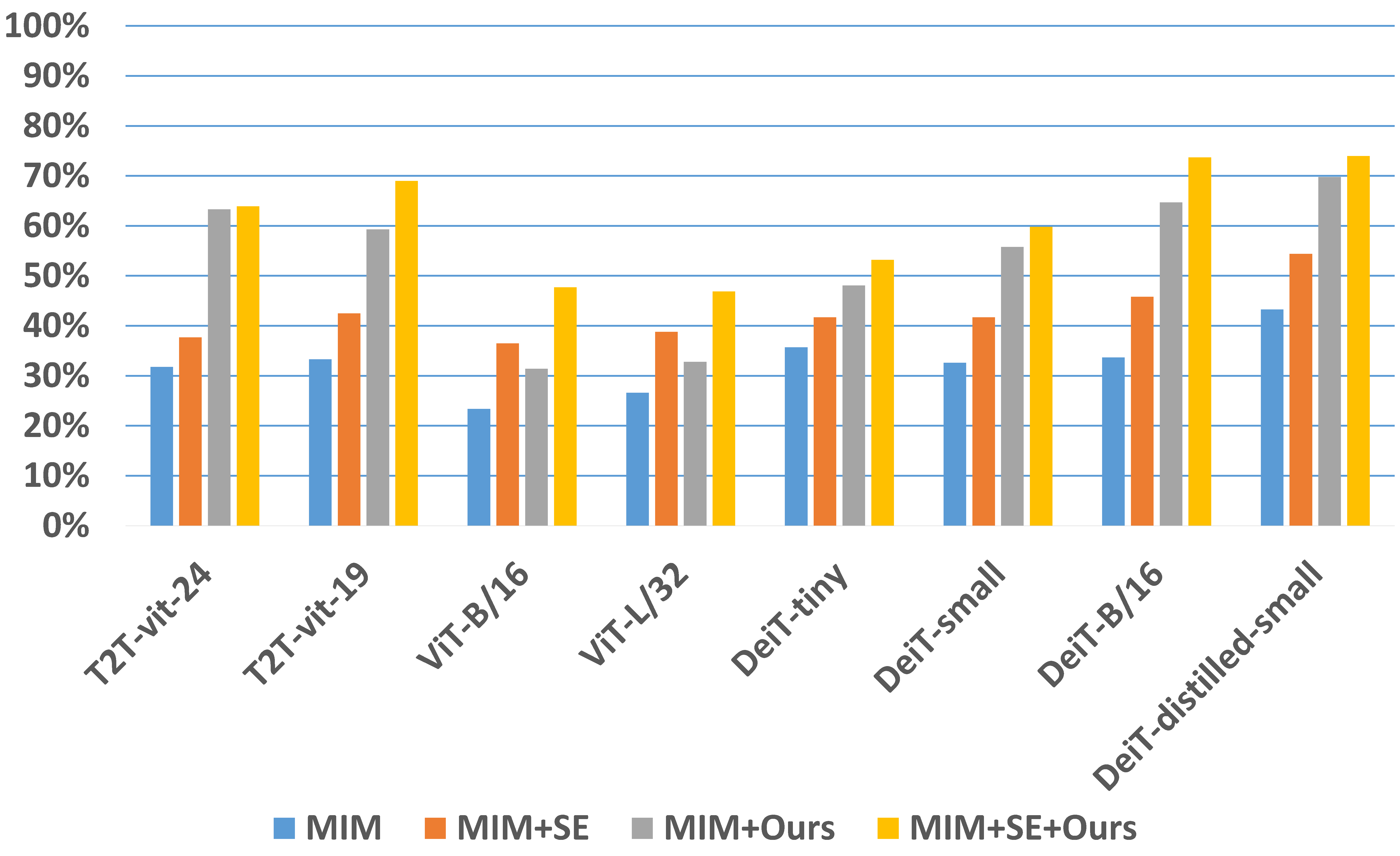} 
	\end{minipage}
    \begin{minipage}[t]{0.33\textwidth} 
			\centering
			\includegraphics[width=\textwidth]{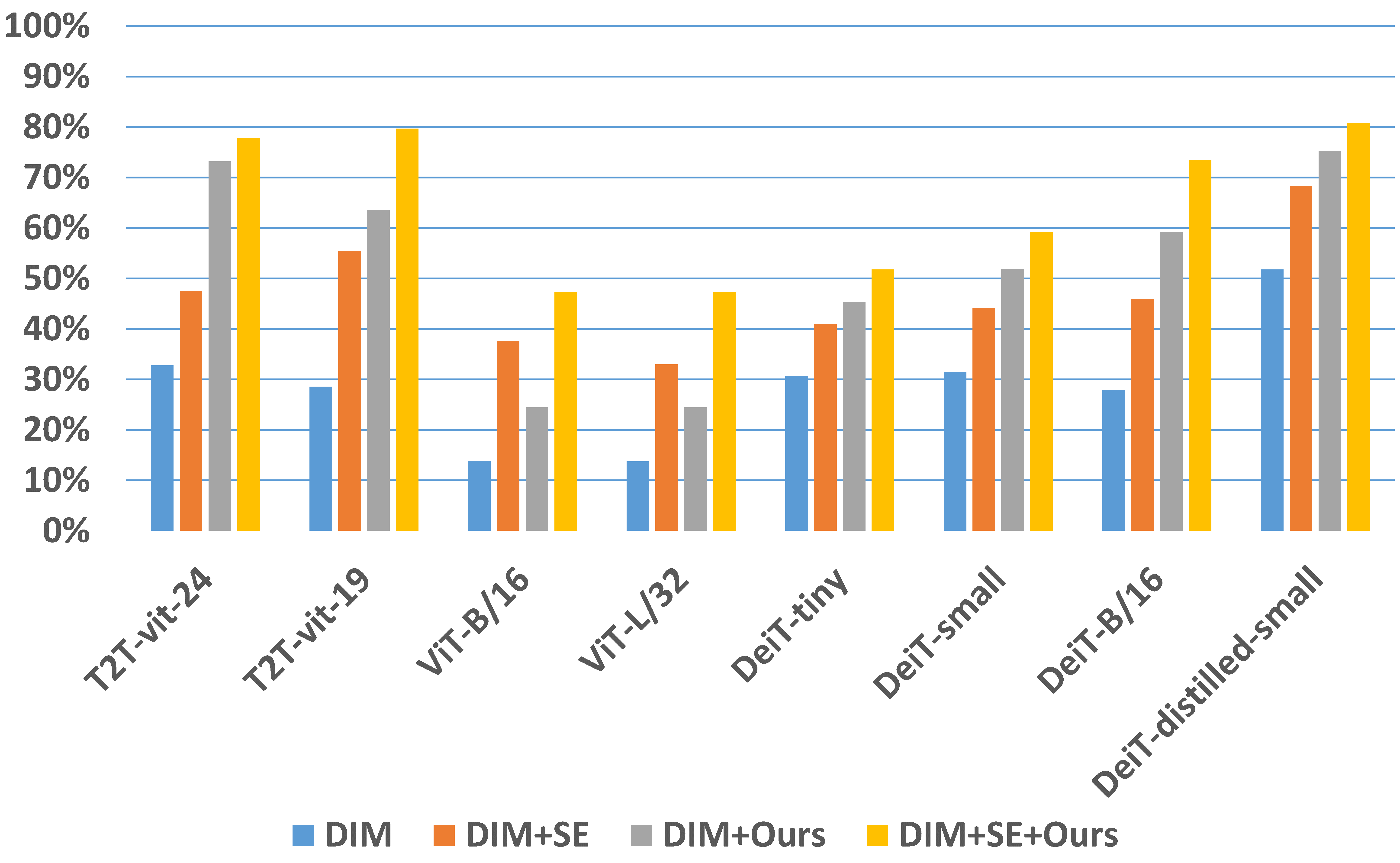} 
	\end{minipage}
	\begin{minipage}[t]{0.33\textwidth} 
			\centering
			\includegraphics[width=\textwidth]{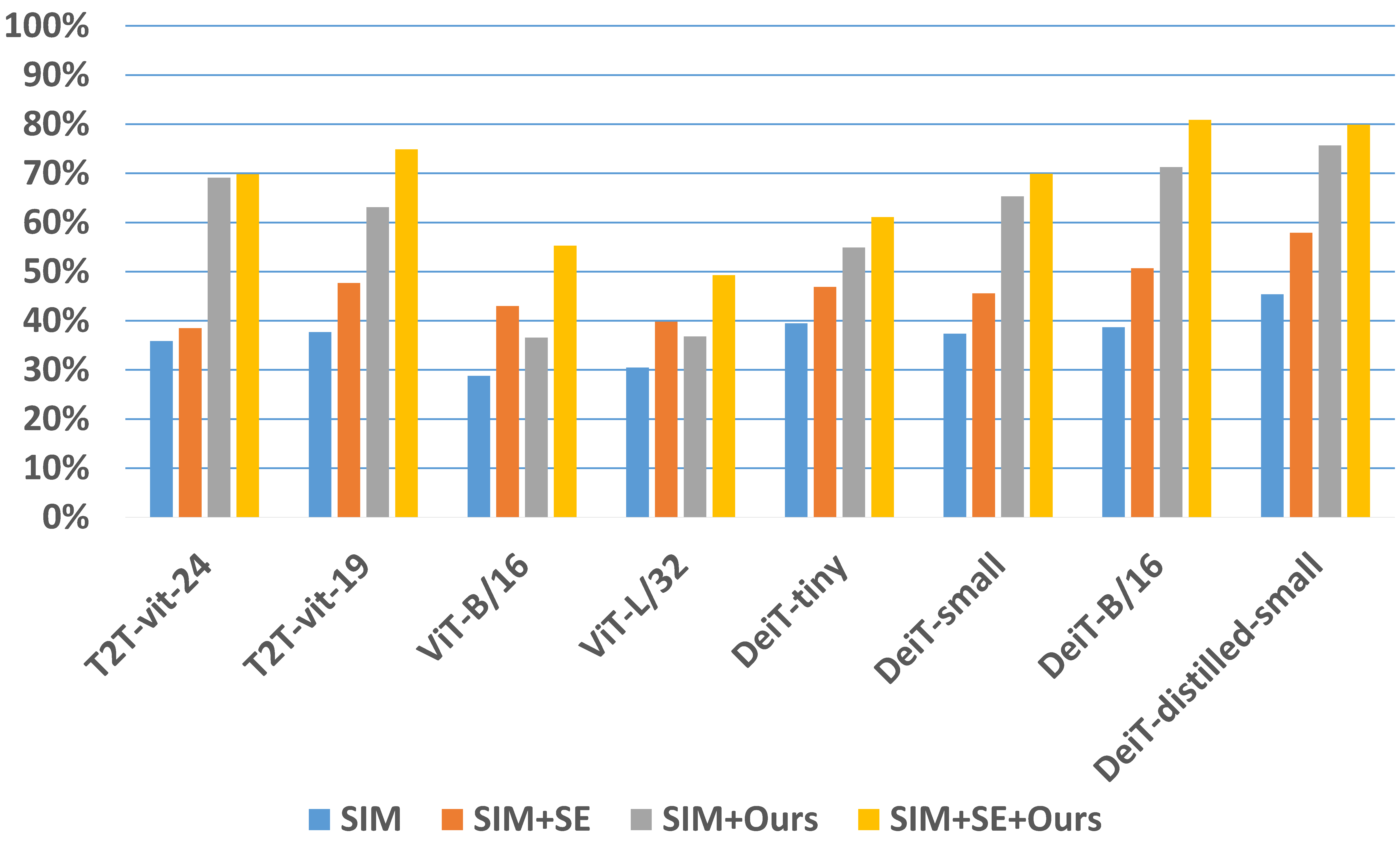} 
	\end{minipage}
    \caption{MIM, DIM, and SIM attack methods are used on 8 ViTs white-box models. The black-box model is chosen as VGG16. The vertical axis indicates the attack success rate on the VGG16 model. The horizontal axis denotes the white-box models. Our method outperforms the SE method on all ViTs white-box models except the ViT white-box series of models but still outperforms the original method. When our SAPR method is combined with the SE method, it exhibits a higher attack success rate.}
    \label{fig:2}
\end{figure*}

\subsection{Probability threshold $\mathcal{P}$}
\begin{figure}[t]
    \centering
    \includegraphics[width=0.95\linewidth,height=0.75\linewidth]{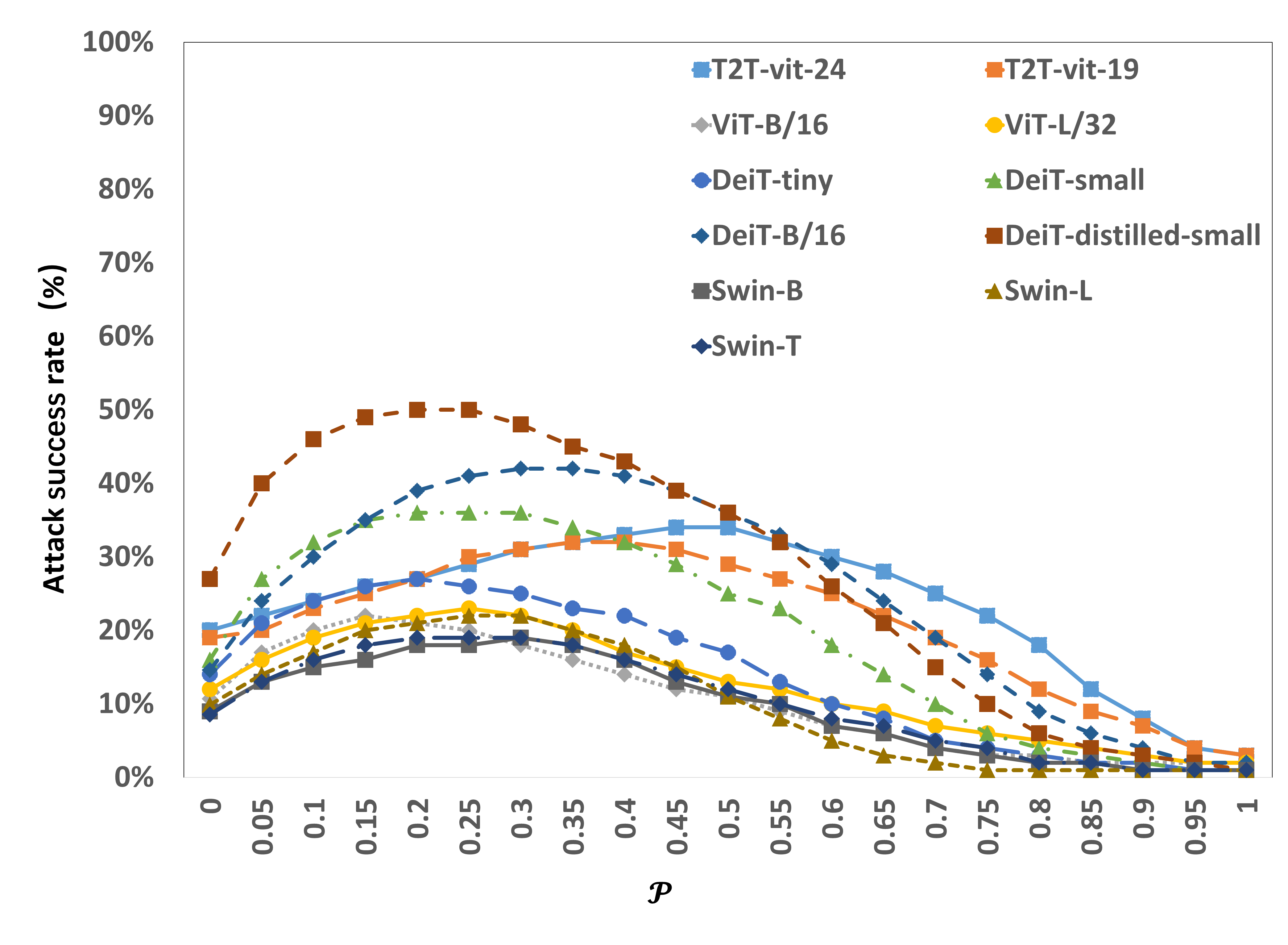}
    \caption{Transferability experiments are performed on 11 ViTs models with probability $\mathcal{P}$ using the MIM method. The horizontal axis represents the value of $\mathcal{P}$ and the vertical axis represents the average attack success rate of the black-box models. Taking values between 0.15-0.45 can enhance the adversarial transferability of ViTs in general.}\label{fig:3}
\end{figure}

The selection of the probability threshold $\mathcal{P}$ will affect the transferability of the adversarial examples generated by ViTs. We use the MIM method on 11 ViTs models. We sampled $\mathcal{P}$ at 0.05 intervals. The experimental results are shown in Fig.\ref{fig:3}, where we report the average attack success rate of the adversarial examples on 60 black-box models with different structures. We found that the effect of $\mathcal{P}$ on the transferability of adversarial samples generated by ViTs tends to increase and then decrease. The extent to which $\mathcal{P}$ affects different ViTs models also differ. When the value of $\mathcal{P}$ is taken between 0.15 and 0.45, it is most beneficial to improve the transferability of the adversarial examples generated by ViTs. Therefore, choosing the suitable probability $\mathcal{P}$ for different ViTs to attack the Attention layer can effectively improve the adversarial transferability. In our experiments, we choose different probability $\mathcal{P}$ values for different models \{T2T-vit-24: 0.5, T2T-vit-19: 0.4, ViT-B-16: 0.15, ViT-L-32: 0.25, DeiT-tiny: 0.2, DeiT-small: 0.3, DeiT-B-16: 0.35, DeiT- distilled-small: 0.25, Swin-B: 0.3, Swin-L: 0.3, Swin-T: 0.3\}

\subsection{Black-box against ViTs}
We generated adversarial examples on four white-box models using the MIM, DIM, and SIM methods and evaluated their transferability on the three black-box ViTs models. The results are shown in Table \ref{table:1}. We report them in more detail in the supplement. From the experimental results, the attack success rate of existing attack methods on black-box ViTs is not yet satisfactory. Our method obtained a higher attack success rate of black-box ViTs when combined with existing methods. We improved the average attack success rate of existing methods on black-box ViTs by \{MIM: 20.08\%, DIM: 15.44\%, SIM: 26.12\%\}. On some black-box ViTs, we have even improved the performance of existing methods by more than double. Our method improves the attack success rate of white-box ViTs on black-box ViTs.

\subsection{Black-box against CNNs and MLPs}
To demonstrate that our proposed method can improve the overall transferability of the adversarial examples generated by ViTs. We not only evaluate the attack success rate of black-box VITs, but we do our best to evaluate black-box models of different architectures. The black-box attack success rate of the adversarial examples generated by ViT is also evaluated on CNN, robust CNN, and MLP using MIM, DIM, and SIM methods. The results are shown in Table \ref{table:1}. From the experimental results, the existing attack methods have low attack success rates on black-box CNNs, robust CNNs, and MLPs due to the structural differences between ViTs and other types of networks. Combining our method with existing methods obtains higher attack success rates for black-box models and even doubles the attack success rate on some black-box models. We improve the average attack success rate of existing methods on black-box CNNs, robust CNNs and MLPs by \{MIM: 16.75\%, DIM: 18.78\%, SIM: 25.00\%\}. It can be learned from all the data in Table \ref{table:1} that our method can improve the attack success rate of white-box ViTs on different structural black-box models, which indicates that our method can overall enhance the transferability of the adversarial examples generated by white-box ViTs making it possible to attack unknown black-box models.

\begin{figure*}[thbp]
    \centering
    \includegraphics[width=\linewidth]{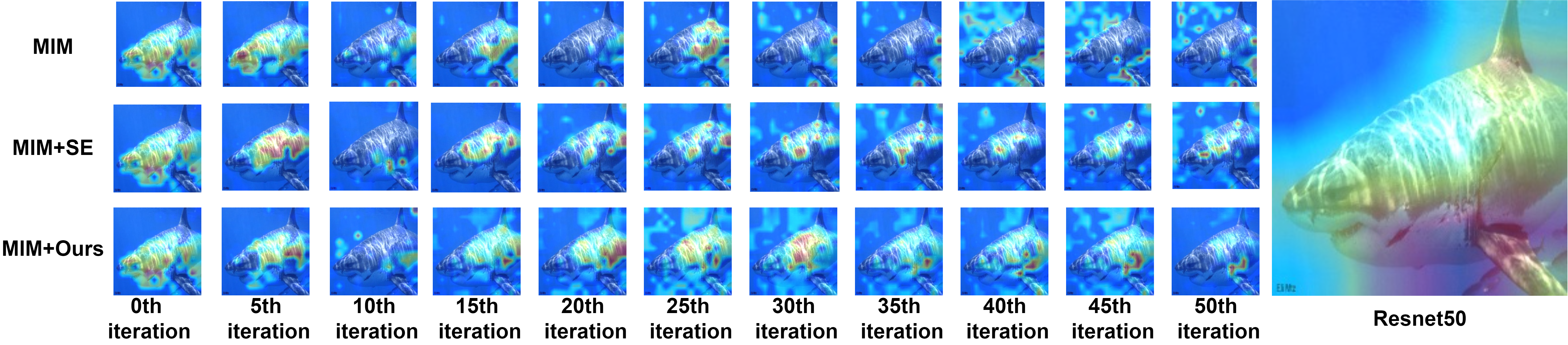}
    \caption{The gradcam images of the MIM, MIM+SE, and MIM+Ours methods during the attack-iterative process are plotted on the white-box T2T-vit-24 model. The last figure shows the gradcam images of the clean image on ResNet50. Our method enables the ViTs model to focus on the object and the global patches. Most importantly, our method always focuses on the object during the iterative attack and is almost consistent with the region of interest of the CNNs model.}\label{fig:4}
\end{figure*}

\subsection{Comparison to state-of-the-art method}
Our research aims to improve the black-box attack ability of white-box ViTs. To the best of our knowledge, the work with the same goal as ours is the recently proposed self-Ensemble (SE) method at the same time is the state-of-the-art method at present. They also propose Token Refine (TR) method to help white-box ViTs to generate adversarial examples with high transferability. However, the TR method requires accurate parameter tuning through the ImageNet dataset to be effective. And we aim to improve the black-box attack ability of white-box ViTs by means of no additional overhead. So here, we do not compare with the TR method. We use the MIM, DIM, and SIM methods to perform experiments on eight white-box models of ViTs. We report the success rate of the attacks on the VGG16 black-box model. The experimental results are shown in Fig.\ref{fig:2}. We report them in more detail in the supplement. By analyzing the experimental results, we obtain the following conclusions. 1. Both our method and the SE method can improve the attack success rate of white-box ViTs on black-box models. 2. On all ViTs white-box models our method outperforms the SE method except for the ViT series white-box models, but still outperforms the original method. 3. Our method can also be easily combined with the SE method, and when the two are combined exhibits a higher success rate of black-box attacks.
\begin{figure}[th]
    \centering
    \includegraphics[width=0.9\linewidth,height=0.9\linewidth]{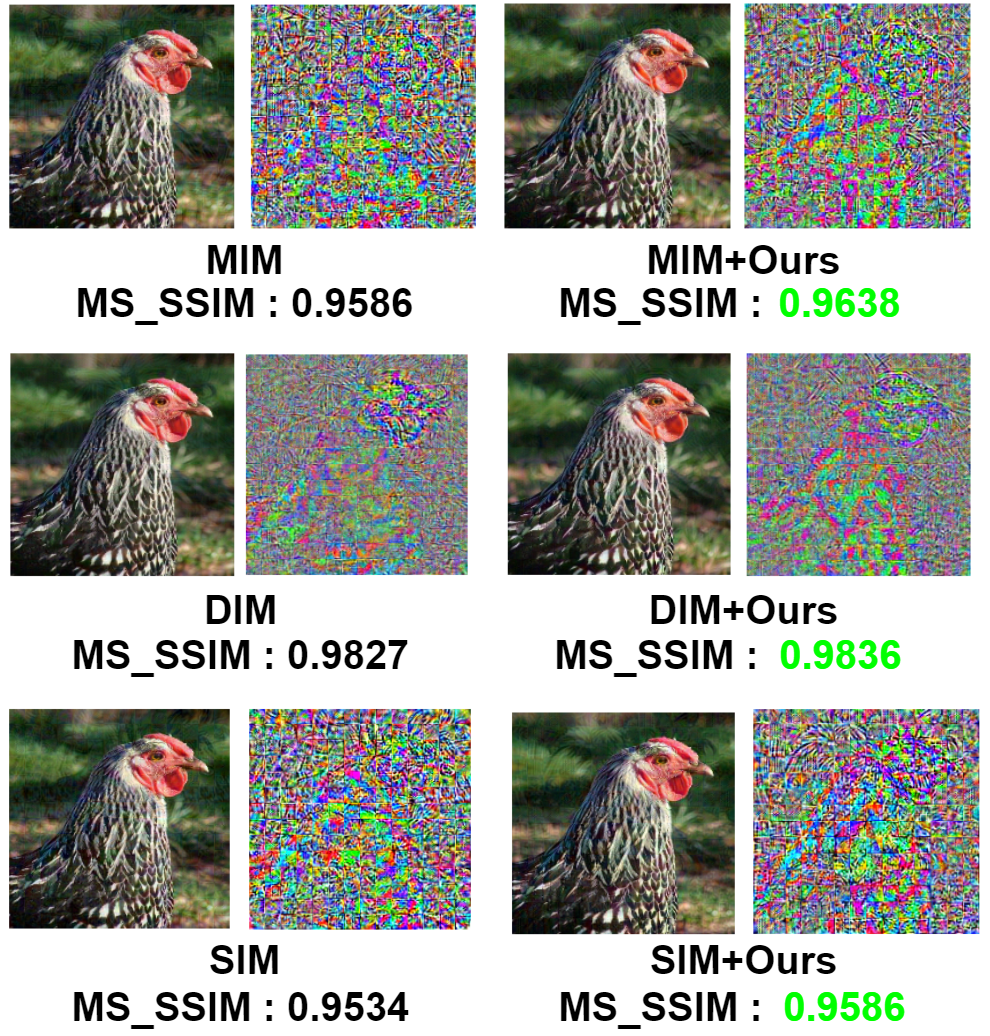}
    \caption{Adversarial examples generated on the DeiT-B/16 model are visualized, and the perturbations generated by our method are more focused on objects and global patches than the original method. By calculating the MS\_SSIM quality score, it is found that our method can generate higher quality adversarial examples.}
    \label{fig:5}
\end{figure}
To explore how our proposed method enhances the transferability of adversarial examples generated by white-box ViTs. We plot the gradcam images of the MIM, MIM+SE, and MIM+Ours methods on the white-box T2T-vit-24 model during the attack-iterative process. The number of iteration rounds is 50. The results are shown in Fig. \ref{fig:4}. The first row shows the gradcam images generated using the MIM method. The second row shows the gradcam images generated using the MIM+SE method. The third row shows the gradcam images generated using the MIM+Ours method. The last figure shows the gradcam image of the clean image on ResNet50. Analysis by results. The original MIM method gradually deviates the region of interest from the object during the iteration and does not focus on the global patches. The SE method can improve the adversarial transferability of adversarial examples because it helps ViTs to keep the region of interest on the object during the iteration. However, the SE method focuses on only a part of the object, unlike our method which can include almost the whole object. Our method can keep the region of interest of ViTs on the object during the iterative process and can focus on the global patches. Compared to the region of interest on ResNet50, the original method deviates from the region of interest on the CNNs model during the iterative attack, which leads to the attack failure. The SE method is successful because the white-box ViTs model can keep the region of interest on the object during the iterative attack and is consistent with the CNNs part of the region of interest. Our method always focuses on the object during the iterative attack and is almost consistent with the region of interest of the CNNs model, so we have a higher success rate of the attack. In summary, our method enables the ViTs model to focus on the global patches and always focus on the objects. Most importantly, we can help ViTs keep the region of interest consistent with CNNs during iterative attacks, thus improving the success rate of attacks on black-box models.

We show in Fig.\ref{fig:5} a visualization of the adversarial examples generated on DeiT-B/16. Our method generates adversarial perturbations more focused on the object than the original method. We calculate the MS\_SSIM values of the adversarial examples with original images. The quality score of the adversarial examples generated using our method is higher than that of the original method (closer to 1 indicates that the image quality is closer to the original image). This indicates that our method generates high transferability adversarial examples and generates images with better quality than the original method. By observing adversarial perturbations, we find that the perturbations generated by our method are mainly concentrated around the object, and we can even see semantic information in them. Concentrated perturbation illustrates that our method helps ViT focus on the object when generating adversarial examples.

\section{Conclusion}
Our work focuses on the transferability of the adversarial samples generated by ViTs on black-box models with different structures. We attack the unique self-attention mechanism in ViTs by restructuring the embedded patches of the input. The restructured embedded patches enable the self-attention mechanism to obtain more diverse patches connections and help ViT keep regions of interest on the object. We also propose an attack method against a unique self-attention mechanism in ViTs, called Self-Attention Patch Restructuring (SAPR). Extensive experiments show that our SAPR method effectively improves the transferability and image quality of the adversarial samples generated by ViTs. Our method improves the transferability of adversarial examples generated by ViTs on black-box models with different structures. Our method is simple to implement but efficient and applicable to any self-attention based network. We make it feasible to attack unknown structural black-box models using white-box ViTs models. We demonstrate that the possibility of this threat arises in ViTs. In the long run, our work will support further research to build more robust deep learning models to defend against our proposed attacks, thus eliminating short-term risks.

\bibliographystyle{named}
\bibliography{ijcai22}

\end{document}